# Surrogate-assisted performance prediction for data-driven knowledge discovery algorithms: application to evolutionary modeling of clinical pathways

*Anastasia A. Funkner*[1], *Aleksey N. Yakovlev*[1,2], *Sergey V. Kovalchuk*[1]
[1]ITMO University, Saint Petersburg, Russia
[2]Almazov National Medical Research Center, Saint Petersburg, Russia
funkner.anastasia@itmo.ru, yakovlev_an@almazovcentre.ru, kovalchuk@itmo.ru

***Abstract.*** The paper proposes and investigates an approach for surrogate-assisted performance prediction of data-driven knowledge discovery algorithms. The approach is based on the identification of surrogate models for prediction of the target algorithm's quality and performance. The proposed approach was implemented and investigated as applied to an evolutionary algorithm for discovering clusters of interpretable clinical pathways in electronic health records of patients with acute coronary syndrome. Several clustering metrics and execution time were used as the target quality and performance metrics respectively. An analytical software prototype based on the proposed approach for the prediction of algorithm characteristics and feature analysis was developed to provide a more interpretable prediction of the target algorithm's performance and quality that can be further used for parameter tuning.

## 1. Introduction

The idea of data-driven knowledge discovery [1] attracts attention in multiple areas where healthcare is not an exception [2]. The area includes diverse and usually non-trivial methods (evolutionary computations, Bayesian methods, fuzzy sets, etc.) applied to the discovery of patterns, functional properties, and dependencies through the analysis of available data. One of the important properties here is the unsupervised nature of the knowledge discovery processes that are applied without or with limited expert involvement. Still, in many cases, the process of discovery is applied to real-world data and faces multiple sources of uncertainty and high complexity of both data and underlying processes and systems. In such cases, the procedure of discovery requires fine-tuning both in terms of performance and in terms of quality. Currently, there exist many works focused on algorithm performance prediction using empirical equations [3] or data-driven models [4,5]. Still, most of them are mainly focused on solutions for computationally intensive numerical algorithms with explicit and measurable quality metrics. On the other hand, knowledge discovery often deals with a) continuous improvement of the obtained solution (knowledge) quality; b) complex assessment of the solution's quality (usually including interpretability, integration with existing domain knowledge, significance, etc.); c) absence of "ideal" solution, especially in case of high uncertainty in the process or system under investigation; d) variation in the algorithm's behavior depending on the particular dataset and problem being solved. Taking this into account, specific data-driven algorithms are required to deal with non-trivial algorithms whose behavior is unknown in advance.

Medicine is one of the areas of high uncertainty and complex knowledge corpus [6]. As a result, knowledge discovery and hybrid (knowledge- and data-driven) [7] algorithms find a broad application in this area. One of the interesting pattern-like structures within this area are clinical pathways (CP) that are used to describe all treatment and patient care processes and include all sorts of events for the treatment of a disease. Often, these clinical pathways are compiled manually with substantial expert involvement [8] or with specific monitoring information systems [9]. Still, due to high complexity and uncertainty in disease development, the task of clinical pathway data-driven discovery is often related to unresolved issues. The list of problems includes the lack of consistency, completeness, and correctness of medical data to be analyzed [10], low coverage of rare cases with CPs [11], weak formalization, and high uncertainty in core medical knowledge [6]. As a result, data-driven, heuristic, and intelligent knowledge discovery algorithms are employed for CP identification and analysis. However, tuning the input parameters for this type of algorithm is time-consuming and may require significant computational powers.

Within this paper, we propose a surrogate-assisted approach for prediction and further optimization of performance and quality of knowledge discovery algorithms with possible automation of their tuning. Within our study, we use previously developed evolutionary algorithms [12,13] for identifying clinical pathways as an example to show the applicability of the proposed approach for real-life clinical problems, linking the results with the solution's interpretability and parameter influence. The remaining part of the paper is organized as follows. Section 2 provides a brief overview of related works in the areas of algorithm quality and performance prediction, CP modeling, and multi-objective optimization. Section 3 reviews the previous authors' works in CP modeling algorithms used as objects for assessing and tuning within the current work. The proposed approach is further described in Section 4. Sections 5 and 6 describe the experimental settings and obtained results, respectively. Section 7 presents a discussion of the possible extensions of the approach investigated in the study. Finally, Section 8 concludes the paper with final remarks.

## 2. Related Works

### 2.1. Metaheuristics and Multi-Objective Optimization

Metaheuristics are a group of methods widely adopted by the research community [14,15]. The concept of a metaheuristic is usually defined as a generic high-level algorithm applied to a wide range of problems (mainly optimization problems) driven by a trial-and-error approach balancing between exploration and exploitation of the search space. Typical examples of metaheuristics are evolutionary algorithms, particle swarm optimization, ant colony optimization algorithms, etc. One of the common issues in the development and application of metaheuristics is proper selection of many algorithms and appropriate configuration of their parameters. The configuration problem is often considered separately in the online (parameter control) and offline (parameter tuning) formats. As a result, there currently is a multitude of methods for solving these problems [16]. The fewFew state-of-the-art examples include such methods as F-Race [17], REVAC [18], CRS-Tuning [19], and many others.

Often, metaheuristic algorithms have high computational complexity as they rely on multiple evaluations of the objective function during the process of optimization. The objective function provides a metric (or multiple metrics) to show how accurately a model approximates the data [20] through trial and error.Often metaheuristic as they rely on multiple evaluation of objective function during optimization process. a(or multiple metrics) through trial-and-error

In this paper, the multi-objective optimization problem (MOOP) is used (a) as an object of modeling and optimization; (b) as a tool to solve an optimization problem. A MOOP never produces a single optimal solution. Moreover, as is usual in a real-life complex task, there are no defined algebraic functions that would determine the relationship between the input data and the output solutions of a MOOP. Therefore, a heuristic approach (HA) is commonly used to solve the problem, for example, the genetic algorithm (GA). As shown in [21,22], HA algorithms are often used with default or most common input parameters. There are several types of numerical methods for solving a MOOP [23]. The first type includes methods where a generalized objective function is constructed with objective functions (e.g. an aggregation function), and solutions are found via the optimization of this one-dimensional function. With this approach, only one solution is optimal. However, the main disadvantage is that one of the objective functions becomes dominant and makes the main contribution to the optimized function at the expense of other objective functions [23]. The second type of numerical methods is based on the evolution of populations (potential solutions of a MOOP). Several indicator functions assess the quality of the solution set, also called an approximation set (see Table 1). It is possible to tune the input parameters of a MOOP solver with the use of a single indicator. However, what can we do when we would like to tune the parameters according to two or more indicator functions? In this paper, we propose an approach to tune a MOOP with many indicators of its solution.

**Table 1:** Tuning input parameters of different MOOP solvers

| Type of MOOP solver | Converting the MOOP to a single objective optimization problem (aggregation of objective functions, optimizing the most important objective, etc.) [23] | Population-based algorithms [23] | |
|---|---|---|---|
| Optimal solution | One optimal (suboptimal) solution | Population of optimal (suboptimal) solutions obtained for the Pareto front | Population of optimal (suboptimal) solutions obtained for the Pareto front |
| Indicators for tuning the input parameters of the MOOP solver | An optimal solution or a number of iterations | One indicator of the approximation set is chosen from cardinality, generation distance, spacing, the hypervolume indicator [24], etc. Custom indicators are possible | More than one indicator |
| Tuning method | Finding the best input parameters of the MOOP solver to minimize the number of iterations or to optimize the solution (grid search, random search, Bayesian optimization, Self-Adaptation, etc.) [25,26] | Finding the best input parameters of the MOOP solver to optimize the chosen indicator of the approximation set (grid search, random search, Bayesian optimization, Self-Adaptation, etc.) [25,26] | With this study, we define the best input parameters if the solutions' indicators belong to its Pareto front. Because of the curse of dimensionality, we suggest using surrogate modeling to predict input parameters for new data |

### 2.2. Surrogate-Assisted Modeling

Recently, multiple ways have emerged where machine learning methods may be used to select an appropriate problem solving method [27] or improve the implemented metaheuristics [28]. One of the more efficient ways is to use machine learning methods to build surrogate models, which are widely used for objective function approximation [29–32].

Surrogate models (often called "metamodels") are commonly used to support working with existing models using approximation of computationally expensive models, with predictions of the outcome or characteristics of the models being applied in complex tasks. The following areas of surrogate modeling can be identified: constrained and non-constrained global optimization [33–35], multi-objectivemultiobjective optimization [30,31], and design space exploration [36,37]. Many studies are aimed to improve existing methods of surrogate modeling [38,39] or to create ensembles of surrogate models [36,38,40–43]. Often surrogate models are parts of evolutionary algorithms acting as fitness functions or individuals of a population [20,30,31,33,44]. The accuracy of surrogate models depends on their structure and the date used to build and train these models. The scientific community has developed intelligent methods for collecting adaptive samples [20,35,42,45], as well as instructions, manuals, and tools to construct surrogate models for the investigated system [42,45,46]. Surrogate models can be classified in different ways [30,31,36,44]. Recently, special environments have been developed to select appropriate surrogate models and tune their parameters automatically [20,47,48]. The authors of [20] created a global surrogate modeling environment with adaptive sampling, in which various types of models are developed using the genetic algorithm and compete for the approximation of iteratively selected data. The authors of [47] developed the COSMOS system, which searches for the optimal model at three levels: the optimal type of model, the optimal type of core, and the optimal values of hyperparameters. The authors of [48] introduced a universal criterion that measures the quality of surrogate models: internal accuracy (using design points), predictive performance (using cross-validation), and roughness penalty.

### 2.3. Algorithm Quality and Performance Prediction

Prediction of an prediction algorithm's quality and performance can reduce the time needed for tuning parameters and improve the quality of the algorithm in a short time. Meta learning is aimed at finding the best algorithm and its parameters for machine learning tasks using previously acquired experience in solving similar problems [49]. However, with any other methods, surrogate modeling or prediction of individual performance parameters is commonly used. The authors of [4] describe an approach for predicting algorithm metrics based on input parameters and descriptive characteristics of the input set, including categorical ones. They also developed a modification of the decision tree and demonstrated their approach by predicting the execution time of the algorithm. The authors of [5] use machine learning methods and genetic algorithms to evaluate the increase in program execution speed using various microarchitectures. In [50], moments of garbage collection are predicted, and the memory profile is estimated using specialized programs. However, the most popular method of parameter tuning is based on genetic algorithms, when the algorithm parameters are individuals of evolution, and the algorithm itself is used as an objective function. In this case, the estimated algorithm is run in each generation, which requires a lot of time and computational resources [51]. Moreover, in this case, it is impossible to obtain a universal model for tuning the estimated algorithm to any data.

### 2.4. Knowledge Discovery in Medicine and Clinical Pathway Modeling

The term "clinical pathway" can be defined in different ways [52,53]. In general, a CP includes methods of treatment, the course of the disease, and other processes occurring with patients. As a rule, a CP is built for a selected group of diagnoses and/or a specific group of patients. There are several approaches for determining clinical pathways. The most common way is to describe CPs manually using medical guidelines and the experience of specialists. Automatic data-driven methods are aimed to discover pathways from real-life processes using hospital data and can be divided into data mining [2] and process mining [54]. In this paper, we use a data-driven evolutionary approach for discovering CPs [12] as an example to demonstrate the surrogate tuning approach for multi-objective optimization of knowledge discovery algorithms.

# 3. Review of Clinical Pathway Modeling Backgrounds

Within this study, we consider the problem ща CP identification as a target algorithm for assessment and optimization. An evolutionary algorithm [12] was developed for clustering and discovering typical CPs that obtain the interpretable structure of typical healthcare processes enabling simulation-based analysis of patient flow [13]. The method was studied using the data of treatment of patients in acute states (using acute coronary syndrome as an example) [12,13] and chronic disease development (using arterial hypertension as an example) [55]. Also, the method was applied to other problem domains, such as predicting purchases by bank clients [56]. This section briefly describes the algorithm, its main features, and functional characteristics. For further details and experimental studies of the algorithm, please refer to the works mentioned earlier in this paragraph.

CPs define the tracks of hospital processes. In our past works, we developed a concept to identify CPs from a hospital's log files and electronic health records. Through the example of patient flow simulation in hospitals, we showed how clinical pathways help to improve the quality of simulation experiments.

---

**Algorithm 1:** Template Discovering through a genetic algorithm and Clustering through them (TDC) [12]

---
1: $S \leftarrow \{S_1, S_2, \dots, S_n\}$ set of sequences of states
2: $krange \leftarrow$ range for cluster numbers to select the best one
3: $alltemplates \leftarrow \mathbf{GeneticAlgorithm}(S)$
4: $kbest \leftarrow \mathbf{ClusterAnalysis}(\mathbf{Kmeans}, alltemplates, krange)$
    ▷ the best number of clusters
5: $representative \leftarrow \mathbf{Kmeans}(alltemplates, kbest).\mathbf{centers}$
6: $clusters \leftarrow \mathbf{Clustering}(S, representative)$
7: **for i from 1 to kbest do**
8:     $cluster \leftarrow clusters_i$
9:     $aligned \leftarrow \mathbf{Align}(cluster, representative_i)$
10:    $\mathbf{ShowCP}(aligned)$

---

In our previous works, we have proposed methods for discovering clinical pathways consisting of several stages including pathways formalization, generation of templates to identify typical CPs, clustering with the Levenshtein distance, and visualization [12]. One of the algorithms for CP identification using a genetic algorithm is the Template Discovering through a genetic algorithm and Clustering through them (TDC). For more details, see Algorithm 1 and Fig. 1. Also, Fig. 2 is an example of clinical pathways for one of the clusters obtained with the TDC method. The cluster presents how the patients move between hospital departments during their hospitalizations. The clusters were interpreted by physicians and used in different projects. For instance, these clusters help to simulate the patient flow through a hospital with regard to its structure and departments.

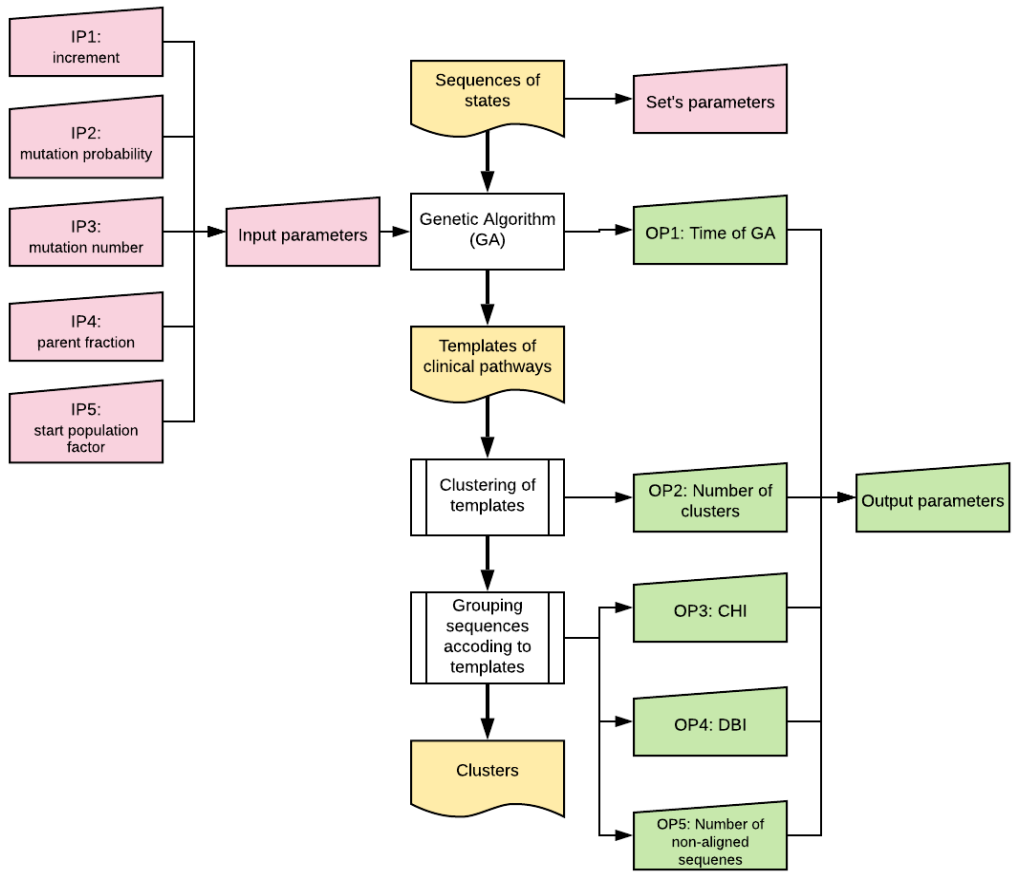

**Figure 1:** Diagram of the TDC method. Clustering metrics: Calinski-Harabaz index (CHI) and Davies–Bouldin index (DBI)

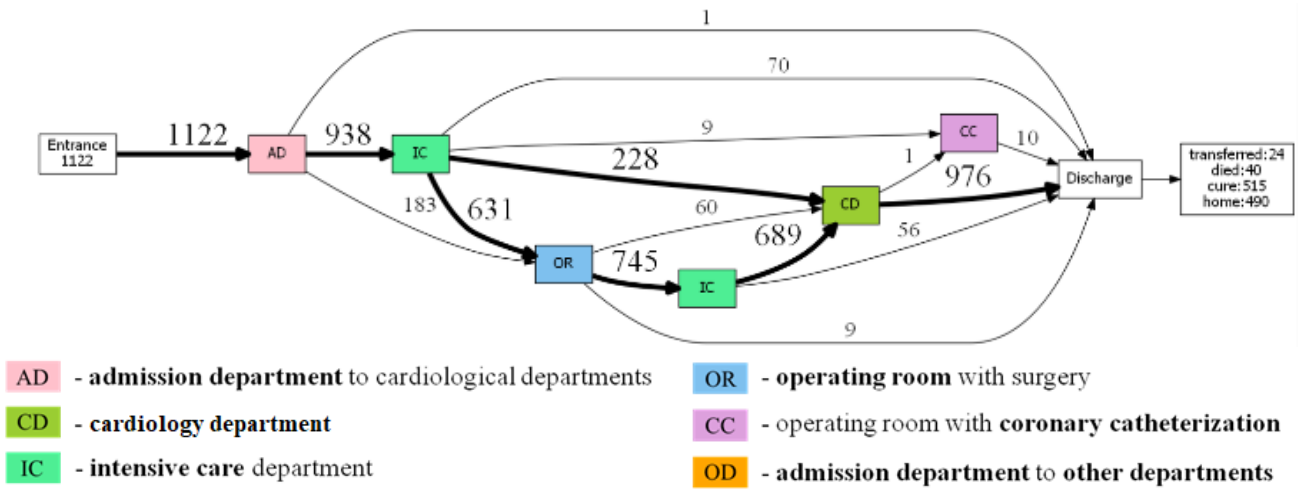

**Figure 2:** Example of typical clinical pathways obtained with TDC

The genetic algorithm (GA) is used in the TDC to generate templates for sequences of states, which can be presented as character strings. In short, a template is a specific state sequence that summarizes all possible states for a group of sequences. The search for such templates is an NPproblem for finding the Shortest Common Supersequence [57]. To solve this problem, we developed the GA that solves the multi-objective optimization problem to find the best templates for a given set of state sequences. The GA consists of the basic steps shown in Fig. 3.

The individuals of populations are character strings. The initial population of candidate solutions is formed randomly using all possible states. An objective function (OF) is a vector function and consists of two components: length and aligning number. The aligning number shows how many sequences of the set fit a template. The change of the Pareto front is defined as a sum of distances from all points of the Pareto front to a zero point $(0, 0)$ in the object space.

The output of the GA depends on many hyperparameters and extremely depends on the structure of an input set of state sequences. Tuning the parameters of the GA with offline methods is a hard task because the launch of the GA needs a significant amount of time. Also, predicting the exact result of the GA with a surrogate model is too hard because the output is a set of sequences, and it crucially depends on the structure of an input set. In this work, we propose several surrogate models to predict the metrics of the output result, namely the time of the GA execution, two clustering metrics, the number of clusters, and the number of non-clustered sequences. Such models allow reducing the time of tuning the GA parameters and exploring these parameters' design space.

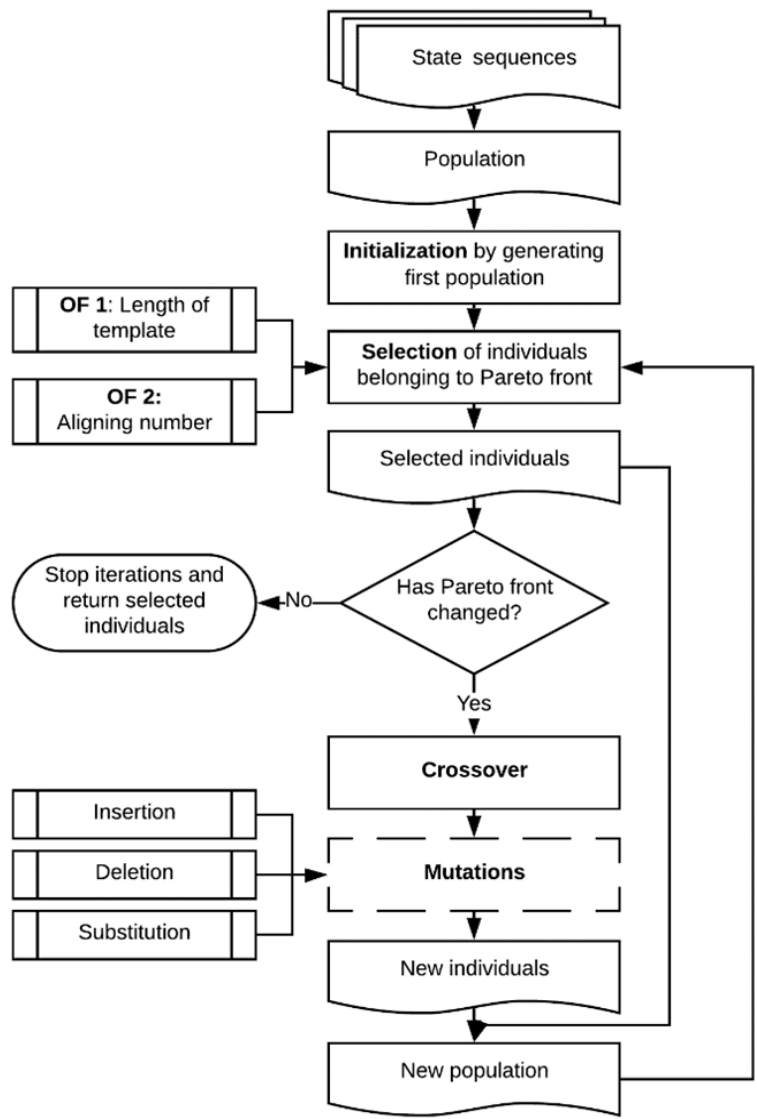

**Figure 3:** Diagram of the genetic algorithm to obtain evolutionary templates
(dotted line indicates probabilistic steps)

# 4. Surrogate-Assisted Prediction and Optimization of Knowledge Discovery Algorithm Characteristics

## 4.1. Proposed Conceptual Basis

Knowledge discovery algorithms can be considered a class of data-driven algorithms with the primary purpose of extension of knowledge either about a particular object or within a problem domain. The available forms of knowledge include functional dependencies between characteristics, robust groups of entities, predictive models, optimization results, etc. The problem of knowledge discovery introduces several other important issues:

1) **Multiple solutions.** Often there is no "perfect solution," and a multitude of forms can be interpreted as a valuable approximation of knowledge. Moreover, a complex landscape of a search space yields no reachable analytical solution. This leads to employing metaheuristics and various numerical techniques to find the solution.
2) **Management of execution time.** In many cases, execution time of a knowledge discovery algorithm is substantial. At the same time, available time is limited due to the available resources or urgent decisions required to be made by a defined deadline.
3) **Domain-specific knowledge quality.** Critical criteria for assessing obtained knowledge are domain-specific: interpretability, relevance, semantic integration, etc. Obtained knowledge should be general and applicable to a particular class of problems within the domain, which leads to the requirements of domain-specific scalability.
4) **General-purpose knowledge quality.** On the other hand, obtained knowledge can have general-purpose quality characteristics: complexity, data coverage, correctness, etc.
5) **Explanatory power and eliminating uncertainty.** The significance of obtained knowledge can be considered in both aspects as absolute or relative values (concerning available expert knowledge). An absolute value is vital for automatic solutions and solving a particular task, while a relative value shows new knowledge added to the domain-specific corpus.

A general idea of surrogate-assisted algorithm prediction and tuning is illustrated in Fig. 4. Regular application of a knowledge discovery algorithm ("direct execution" in Fig. 4) usually includes algorithm tuning, execution, and interpretation of data. A key issue here is the balance between execution time and quality of output knowledge. Moreover, many tuning algorithms also rely on the execution history and require a lot of runs during the parameter optimization process. As a solution to overcome this issue, surrogate assisted tuning is aimed at eliminating or reducing the actual execution of an algorithm by replacing it with a surrogate model/algorithm and predicting the output data, characteristics of the execution process, or obtained knowledge ("surrogate-assisted" in Fig. 4). A surrogate model may use the characteristics of input data, algorithm (algorithm parameters), and execution environment (both software and hardware) as parameters for predicting algorithm characteristics. One of the challenging issues is identification of proper surrogate models ("surrogate training" in Fig. 4figure). In the case of data-driven algorithms, this issue becomes even more complicated, as the domain-specific diversity in data may influence the algorithm performance significantly. Moreover, structuring this diversity is often the goal (and the output) of the algorithm. Within the current study, we focus on the analysis of the influence of various data characteristics (both input and output of the target algorithm) of performance prediction using surrogate models.

**Figure 4:** Surrogate-assisted algorithm tuning

In the general case, there are two main groups of characteristics: performance (obtained directly after the execution of the algorithm) and quality (obtained after the interpretation of the execution results). Mainly, the characteristics of these two groups have an inverse relationship: longer execution time leads to a higher quality of obtained knowledge. At the same time, due to the existing limitations, surrogate-assisted tuning can be an essential technique for reaching a higher quality of the results within a limited time (e.g., in a deadline-driven approach [58]). Knowledge discovery algorithms introduce a higher complexity of quality assessment as it involves multiple criteria and specificity of the domain knowledge used for interpretation. Within the presented study, we consider an MOOP-based approach for an algorithm's assessment and tuning, which may be used in various knowledge discovery tasks.

### 4.2. Preliminaries

This section introduces the main definitions used in the remainder of the paper.

**Definition 1 (sequence of states).** Let $A$ be a finite set of all possible states. A sequence of states is an ordered collection of states: $s = <a_1, a_2, ..., a_k>, a_i \in A$. Let $S$ be a set of sequences of states. The set $S$ can be described with parameters $P_S$, also called set's parameters.

**Definition 2 (evolutionary algorithm).** Let $E$ be an evolutionary algorithm with hyperparameters $P_E$, also called input parameters. Let $E$ solve the multi-objective optimization problem (MOOP).

**Definition 3 (solution).** Let $S$ be solutions of the MOOP obtained with the evolutionary algorithm $E$: $E(S, P_E) = \hat{S}$. $\hat{S}$ is an approximation set of Pareto optimal solutions for the MOOP, which is also called the Pareto front in the object space. The solution $\hat{S}$ can be described with parameters $P_{\hat{S}}$, also called output parameters.

**Definition 4 (surrogate model).** Let the algorithm $M$ be a surrogate model if $M(P_{input}) = P_{output} + \varepsilon_M$ where $\varepsilon_M$ is an error of $M$ where $P_{input}$ includes hyperparameters $P_E$ or/and set's parameters $P_S$.

Surrogate models are developed to imitate output characteristics of a method. We propose two approaches to surrogate modeling. With the first approach, a surrogate model is built for a specific set to find relations between input and output parameters of a method. With the second approach, a surrogate model is built for all available data, and the set's parameters are used as input for surrogate models. Moreover, we propose to build ensembles of models obtained with the first approach.

### 4.3. Surrogate Models for Each Set of Sequences

Let $S_{all}$ be a set of sets of sequences: $S_{all} = \{S_1, S_2, \ldots, S_n\}$. Surrogate models for each set of sequences are built as shown below:

$$\mathbf{M}_{each} = \{\mathbf{M}_i : \mathbf{M}_i(P_E) = P_{\widehat{S}_i} + \varepsilon_{M_{1i}} \; for \; S_i \in S_{all}\}.$$

It is expected that machine learning models (MLMs) are used as surrogate models. MLMs are usually trained with a training set first and then are checked with a test set to calculate the accuracy or other metrics of MLMs. The models $\mathbf{M}_{each}$ are trained with a specific set of sequences and adapt to its features.

### 4.4. General Surrogate Model for Sets of Sequences

A general surrogate model is trained under sets' parameters $P_S$ and algorithm's parameters $P_E$:

$$\mathbf{M}_{gen}(P_E, P_S) = P_{\widehat{S}} + \varepsilon_{gen}.$$

### 4.5. Ensembles of Surrogate Models

Ensembles aggregate the results of several base models to improve the accuracy of these models' output [59]. There are two common aggregation functions: voting (used for classification) and weighted averaging (used for regression). In this section, we present two approaches to build ensembles with averaging. With the first approach, the outputs of all surrogate models for each set are just averaged. With the second approach, the most appropriate models are selected first, and then the average output is calculated.

*4.5.1 Average ensemble of surrogate models.* The average ensemble of surrogate models is based on the surrogate models for each set. For any input set, this ensemble gives the average solution despite the parameters of the input set:

$$\mathbf{M}_{aver} = \frac{1}{n} \sum_{i=1}^{n} \mathbf{M}_i(P_E).$$

*4.5.2 K-nearest neighbors ensemble of surrogate models.* The k-nearest neighbors ensemble considers the parameters of the input set. For a new set $S_{new}$ $k$ the most similar sets of sequences are selected by comparing the parameters of $S_{new}$ with the parameters $P_{S_{all}}$ used for building the $\mathbf{M}_{each}$ where $P_{S_{all}} = \{P_{S_i} for \; S_i \in S_{all}\}$.

$$\mathbf{M}_{neig} = \frac{1}{k} \sum_{i=1}^{k} \mathbf{M}_{j_i}(P_E),$$

where $j_1, j_2, \ldots, j_k$ are the indices of k-nearest sets of sequences to the new set $S_{new}$.

## 5. Clinical Pathway Identification: Acute Coronary Syndrome

### 5.1. Data Description

In this research, we used a set of 3434 electronic health records (EHRs) of ACS patients admitted to the Almazov National Medical Research Centre (Almazov Centre)[1] during 2010–2015. Patients with acute coronary syndrome (ACS) usually stay in various departments during their treatment, e.g. admission department, regular care department, surgery room, and intensive care department, and can move between these departments because of their health condition and hospital schedule. Hospital departments are considered as states in this experiment. Each EHR is associated with a sequence of patient movements between departments. There are 229 unique sequences of patient movement, so the obtained sequences are different. We used these 229 sequences to generate new sets for the experiment. A set's generator produces a new set with given templates and the probability of mutations.

---
[1] http://www.almazovcentre.ru/?lang=en

Using these 229 sequences of the initial set of patients, we generated different sets of sequences for experiments. Besides the initial set, there are clusters of similar sequences that were derived with our method for identification and clustering of clinical pathways [12]. These clusters were obtained from the EHRs of ACS patients and were interpreted by physicians. For example, Cluster #5 contains patients whose treatment strategy agrees with clinical recommendations in the best way. Cluster #8 presents people with myocardial infarction who were delivered by an ambulance in a state of cardiogenic shock or clinical death. Moreover, the generated sets include mixes of clusters, two separate sets with short and long sequences, three randomly generated sets, and 24 template sets, which were generated with mutated typical templates. Template sets can be obtained from a different number of templates (from 1 to 10). The list of sets and statistical parameters of sequence length is presented in Table 2.

For each set, the TDC method was run 3125 times (five random values for five GA parameters) on an irregular grid of parameters of the GA. The results of these launches were used as samples to train surrogate models. Also, the data for the surrogate models was divided into training and testing sets with a 70:30 split.

**5.2. Parameters**

*5.2.1 Parameters of sets.* For some surrogate models, the parameters of sets ($P_S$) of sequences are used as input data. These parameters can be divided into two classes: length parameters of sequences and frequencies of states. The length parameters are minimum and maximum lengths, a median, and a standard deviation of lengths among sequences of a set. Furthermore, there is a parameter of number of length outliers calculated with the interquartile range. The frequencies of n-grams of states are also used as input data. Also, there is a parameter of unique sequences in a set. In this experiment, we used 1-grams and 2-grams to describe the sequences of sets.

**Table 2:** Parameters of sets for experiments

| Set name | Min of length | Max of length | Median of length | St. dev. of length | Outliers of length | Unique sequences (cardinality) |
|---|---|---|---|---|---|---|
| Initial set | 1 | 12 | 5 | 1.59 | 243 | 229 |
| Cluster 1 | 6 | 10 | 9 | 0.78 | 4 | 51 |
| Cluster 2 | 5 | 6 | 5 | 0.5 | 0 | 7 |
| Cluster 3 | 1 | 5 | 3 | 0.97 | 11 | 42 |
| Cluster 4 | 6 | 9 | 8 | 0.67 | 3 | 69 |
| Cluster 5 | 5 | 7 | 7 | 0.75 | 0 | 6 |
| Cluster 6 | 4 | 8 | 7 | 0.72 | 27 | 86 |
| Cluster 7 | 9 | 13 | 11 | 0.99 | 7 | 40 |
| Cluster 8 | 4 | 5 | 4 | 0.48 | 0 | 11 |
| Cluster 9 | 4 | 6 | 5 | 0.74 | 0 | 37 |
| Clusters 2, 5 | 5 | 7 | 6 | 0.77 | 0 | 13 |
| Clusters 3, 4, 9 | 1 | 9 | 6 | 2.19 | 3 | 148 |
| Clusters 4, 9 | 6 | 9 | 8 | 0.67 | 3 | 73 |
| Clusters 5, 9 | 4 | 7 | 5 | 0.86 | 17 | 43 |
| Short sequences | 1 | 6 | 5 | 1.2 | 10 | 98 |
| Long sequences | 7 | 12 | 8 | 1.48 | 12 | 131 |
| Random set 1 | 1 | 12 | 7 | 3.48 | 0 | 211 |
| Random set 2 | 1 | 12 | 6 | 3.52 | 0 | 206 |
| Random set 3 | 1 | 12 | 7 | 3.39 | 0 | 215 |
| Template set 1 (1 template) | 4 | 10 | 8 | 1.16 | 24 | 119 |
| Template set 2 (1 template) | 10 | 16 | 14 | 1.21 | 25 | 141 |
| Template set 3 (1 template) | 5 | 11 | 9 | 1.17 | 18 | 111 |
| Template set 4 (2 templates) | 7 | 13 | 10 | 1.26 | 12 | 144 |
| Template set 5 (2 templates) | 4 | 16 | 8 | 2.97 | 3 | 165 |
| Template set 6 (3 templates) | 4 | 15 | 9 | 2.76 | 1 | 178 |
| Template set 7 (3 templates) | 5 | 18 | 8 | 3.37 | 13 | 163 |
| Template set 8 (3 templates) | 8 | 19 | 15 | 3.01 | 0 | 190 |
| Template set 9 (5 templates) | 4 | 15 | 8 | 1.98 | 30 | 174 |
| Template set 10 (5 templates) | 6 | 18 | 14 | 2.56 | 20 | 206 |
| Template set 11 (5 templates) | 3 | 17 | 9 | 3.01 | 13 | 189 |
| Template set 12 (6 templates) | 5 | 17 | 11 | 2.93 | 4 | 202 |
| Template set 13 (6 templates) | 5 | 16 | 9 | 1.94 | 15 | 196 |
| Template set 14 (7 templates) | 5 | 16 | 10 | 2.35 | 16 | 206 |
| Template set 15 (7 templates) | 4 | 19 | 10.5 | 3.81 | 2 | 204 |
| Template set 16 (8 templates) | 5 | 18 | 10 | 2.83 | 25 | 219 |
| Template set 17 (8 templates) | 4 | 17 | 9 | 3.23 | 24 | 180 |
| Template set 18 (8 templates) | 5 | 18 | 9 | 3.48 | 4 | 203 |
| Template set 19 (9 templates) | 3 | 17 | 9 | 2.83 | 16 | 220 |
| Template set 20 (9 templates) | 5 | 17 | 9 | 2.84 | 14 | 204 |
| Template set 21 (9 templates) | 4 | 18 | 11 | 3.86 | 0 | 213 |
| Template set 22 (10 templates) | 4 | 17 | 9 | 2.36 | 15 | 205 |
| Template set 23 (10 templates) | 5 | 17 | 9 | 2.59 | 9 | 219 |
| Template set 24 (10 templates) | 4 | 13 | 8 | 1.81 | 19 | 189 |

*5.2.2 Parameters of an evolutionary algorithm.* The TDC method includes a genetic algorithm. This genetic algorithm is built according to the scheme shown in Fig.3 and has the following input parameters ($P_E$):

- an **increment** means how many times the length of the template may exceed the longest sequence in the population;
- a **mutation probability** is a tuple of three probabilities for each type of mutation, the sum of probabilities equal to or less than one;
- a **mutation number** is the maximum possible number of mutations for a sequence;
- a **parent fraction** is the share of parents in the population;
- a **start population factor** means how many times the population size exceeds the size of the initial set of sequences.

*5.2.3 Output parameters.* The output of the GA is a set of best solutions. The representative templates are selected among the best solutions as centers after clustering the approximation set of the Pareto front with k-means methods. The representative templates are used to divide the initial input set into clusters, which can be visualized with graphs of typical clinical pathways, as shown in Fig. 2. The developed surrogate models predict the parameters ($P_{\hat{S}}$) of launches of the TDC method and the parameters of the clusters:

- **time** of GA execution in seconds;
- **number of clusters** obtained with represented templates;
- **Calinski-Harabaz index** clustering metric, also known as the variance ratio criterion, first local maximum of this index shows the optimal number of clusters [60];
- **Davies–Bouldin index** clustering metric, its minimum value shows the optimal number of clusters [61];
- **number of non-clustered sequences**, if there are outliers among the initial sequences of the input set, it is possible that some sequences will not be clustered.

### 5.3. Surrogate Models for Each Set of Sequences

We selected random forest regression as a surrogate model ($\mathbf{M_{each}}$) because of its simple interpretation and short training time. A separate regression model was built to predict each output parameter. The following hyperparameters of the random forest models were selected with the grid search: number of decision trees, maximum possible depth of decision trees, minimum of samples for node splitting. The mean absolute percentage error (MAPE) is used to validate the regression models:

$$MAPE = \frac{100\%}{n} \sum_i \frac{|y_{true_i} - y_{pred_i}|}{y_{true_i}},$$

where $y_{true}$ are true values and $y_{pred}$ are predicted values calculated with a regression model for $i$ sample from the validation set.

### 5.4. General Surrogate Models for Sets of Sequences

In the previous section, a separate surrogate model ($\mathbf{M_{each}}$) was built for each input set. This approach is appropriate if it is necessary to explore the design space of the parameters of TDC methods. However, this approach does not allow tuning of the parameters of TDC if a new set of sequences is provided. As a result, a general model ($\mathbf{M_{gen}}$) is a model that is trained in the data of all sets and considers the parameters of input sets ($P_S$) and the parameters of the GA ($P_E$). Two models were tested to solve this problem: a neural network (a multilayer perceptron) and separate random forest regressors for each output parameter.

### 5.5. Ensembles of Surrogate Models

An ensemble of separate surrogate models can be built if it is too expensive to train a general surrogate model or if there is not enough information about the set of sequences. Still, it is necessary to explore the design space of the model. Both an average ensemble of surrogate models ($\mathbf{M_{aver}}$) and a k-nearest neighbors' ensemble of surrogate models ($M_{neig}$) were built using RF models, as mentioned in Section 4.3.

## 6. Results and Analysis

For the surrogate models of each set ($M_{each}$), Fig. 4 and 5 show the biplots of real and predicted values of output parameters. In a perfect case, points on a biplot should lay on a diagonal line. As can be seen, the biplots for predicted values of training and tests look similar, so there is no overfitting of the models. In Fig. 4 and 5, the initial set contains all possible sequences of states from real data, so the biplots for it contain more different points than for cluster #5. The more diverse the sequences in the set, the more diverse the results obtained after applying the GA. It is particularly evident in the example of the Davies-Bouldin index. Moreover, the initial set contains many more outliers, which are the cause of the abnormal results of the TDC launches, which is seen in the example of the Calinski-Harabaz index and the number of clusters.

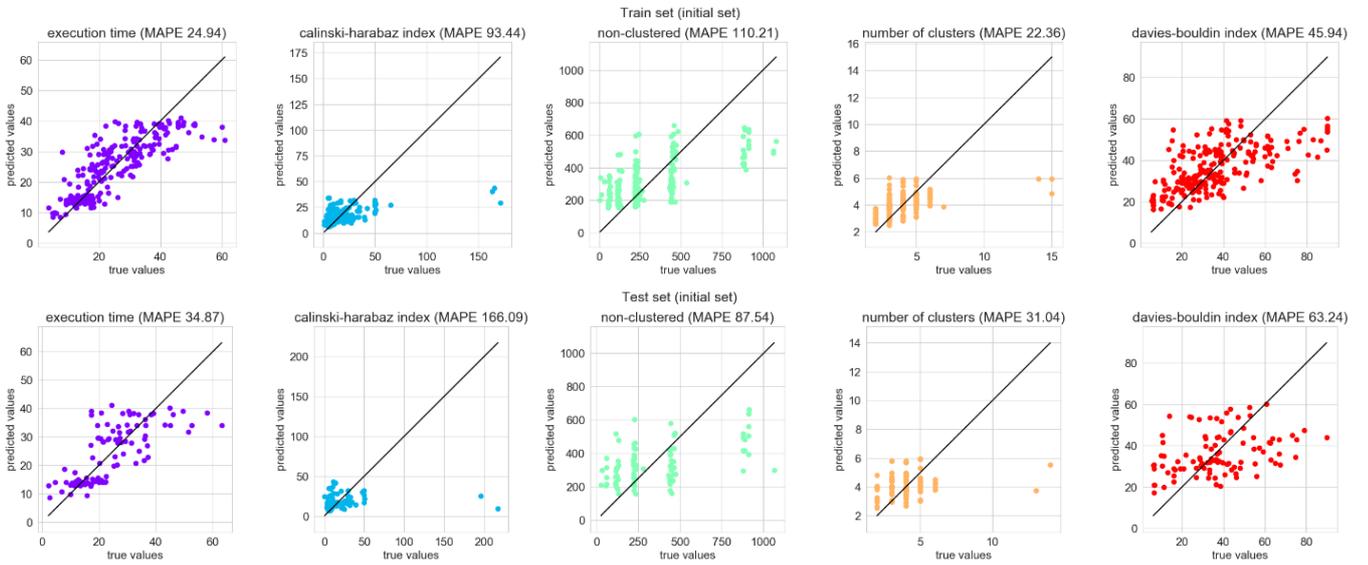

**Figure 4:** Predicted output parameters for the initial set with surrogate models for each set

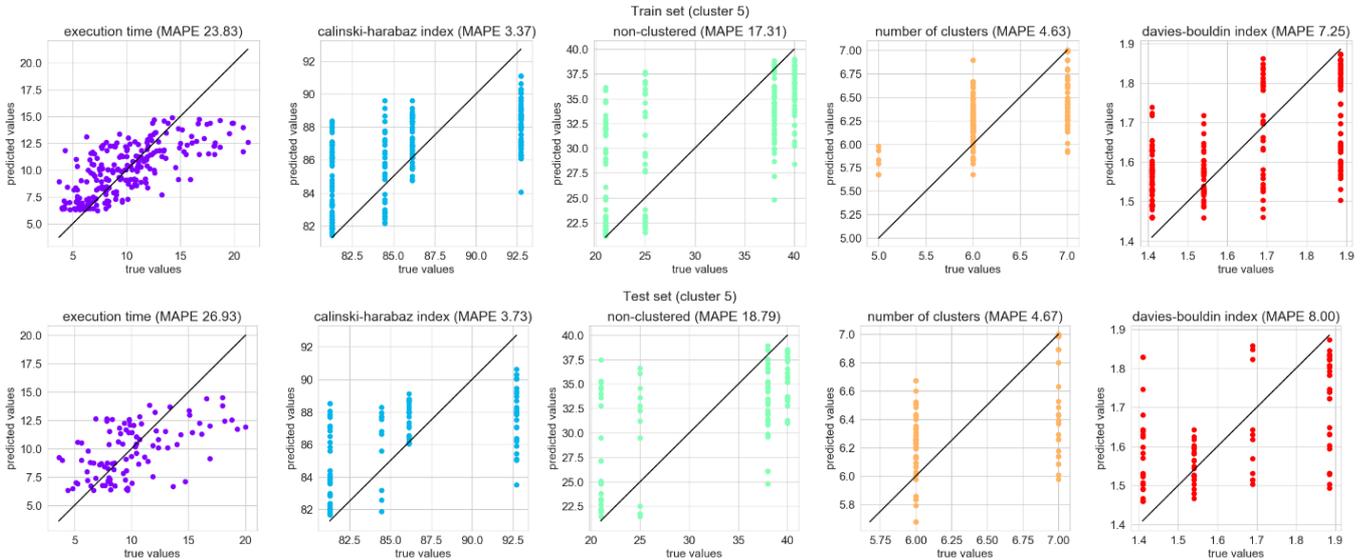

**Figure 5:** Predicted output parameters for cluster #5 with surrogate models for each set

For the general models ($M_{gen}$), Fig. 6 and 7 show the difference in MAPE of both the neural network (NN) and the random forest (RF) models. In Fig. 7, it is clearly seen that, despite the long search through the parameters, it is not possible to find such parameters for the NN so that it could describe all the patterns of the input data. Using the number of clusters and the Davies-Bouldin index, it is clear that the NN could not learn to determine a large group of parameters and gave them a constant response, which is depicted as a straight line on the biplots. Also, the NN determines the longest possible execution time for the launches (about 75–80 seconds), which also lowers the prediction accuracy. The RF model was able to distinguish subgroups in the input data, as can be seen in the Calinski-Harabaz index (Fig. 8) and the number of non-clustered sequences. The RF model is better in many indicators: MAPEs are less for all output parameters and the time of training is much less. The training time of RF is 7.18 sec, whereas the time of NN is 146 sec. The parameters of both models were selected with the random search, and the time of the random search was 18 min 34 sec for RF and 1 h 12 min 46 sec for NN.

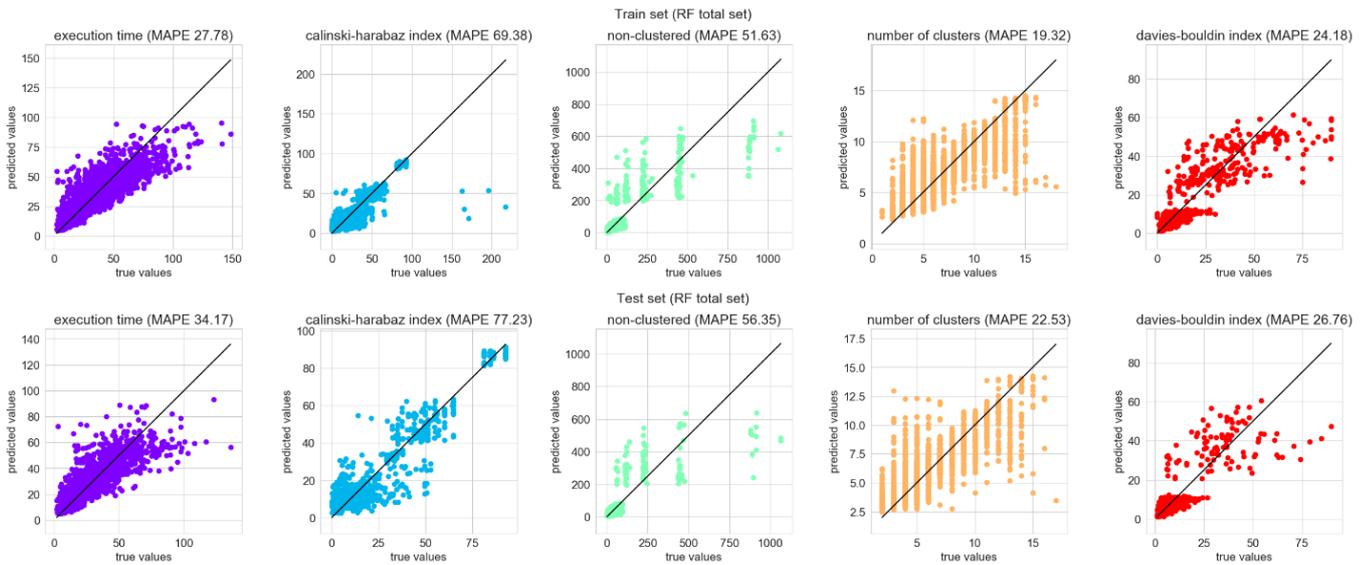

**Figure 6:** Predicted output parameters for the total set with the random forest model (RF)

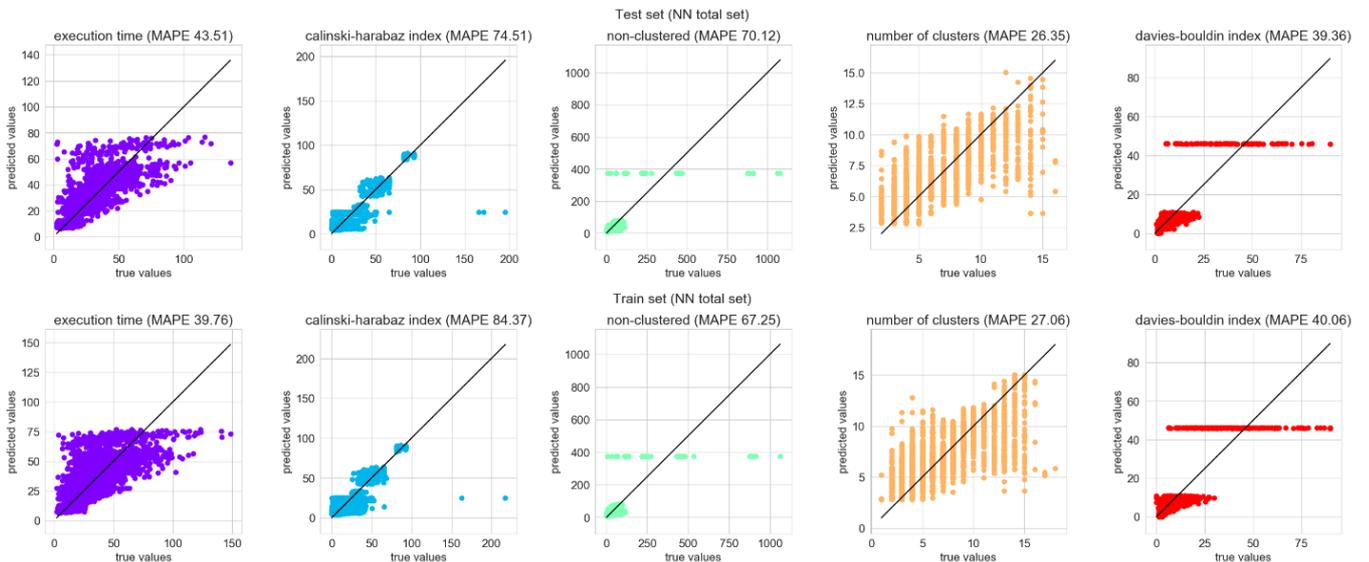

**Figure 7:** Predicted output parameters for the total set with the neural network (NN)

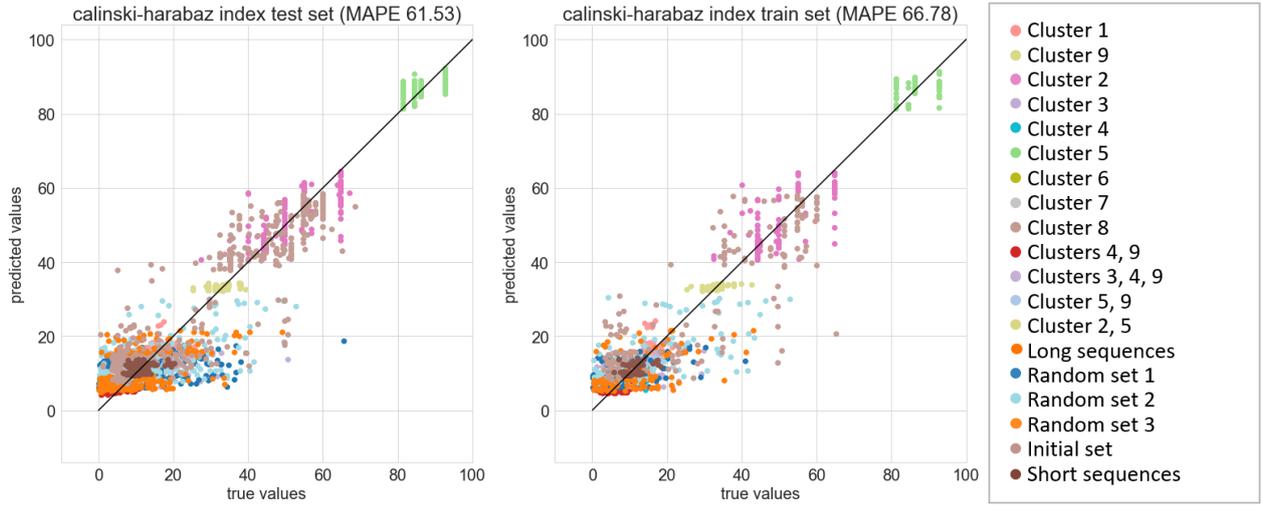

**Figure 8:** Groups of launches in the biplot of the Calinski-Harabaz index of the general model RF for non-template sets

**Table 3:** The average MAPE for surrogate models and ensembles

| Type of surrogate model | Calinski-Harabaz index | Davies-Bouldin index | Execution time | Non-clustered | Number of clusters |
|---|---|---|---|---|---|
| Separate models | 82.68 | 28.19 | 34.31 | 58.08 | 23.04 |
| General model (RF) | 61.53 | 24.19 | 27.8 | 51.7 | 19.37 |
| General model (NN) | 18305410 | 11727122 | 35.09 | 69.91 | 25.59 |
| Average ensemble | 255.21 | 120.03 | 82.32 | 157.44 | 48.53 |
| K-nearest neighbors' ensemble | 103.74 | 46.7 | 47.36 | 91.23 | 47.12 |

Table 3 shows the average MAPE for surrogate models and their ensembles. The general model has the best result for all parameters. According to Table 3, the ensembles have the lowest MAPEs. However, if the surrogate models for each set ($M_{each}$) are built, ensembles do not need time to train. Though the ensemble $M_{neig}$ uses the method of k-neighbors classification, the training of this classification is memorizing all samples of a training set. The k-neighbors classification does not perform any calculations during the training [62].

a)

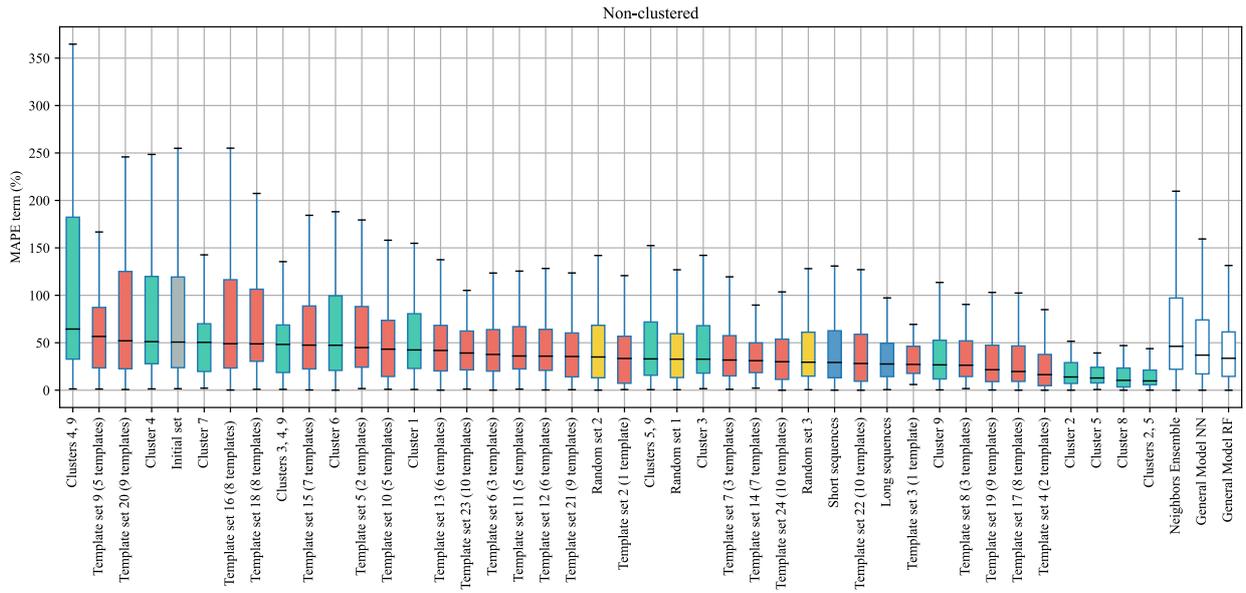

b)

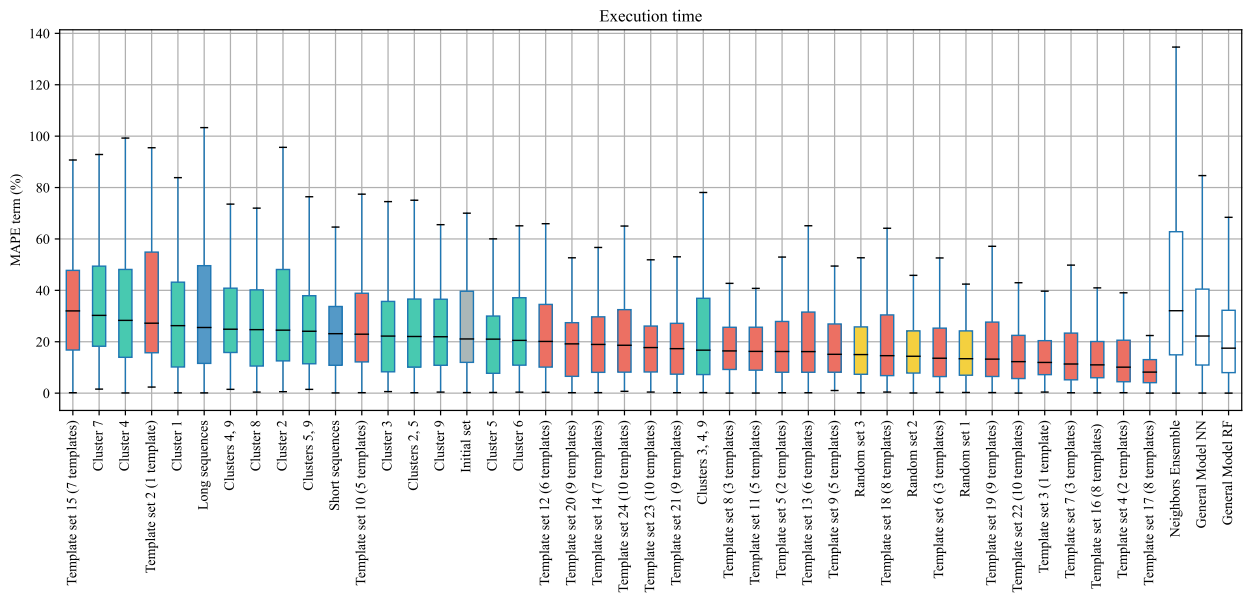

c)

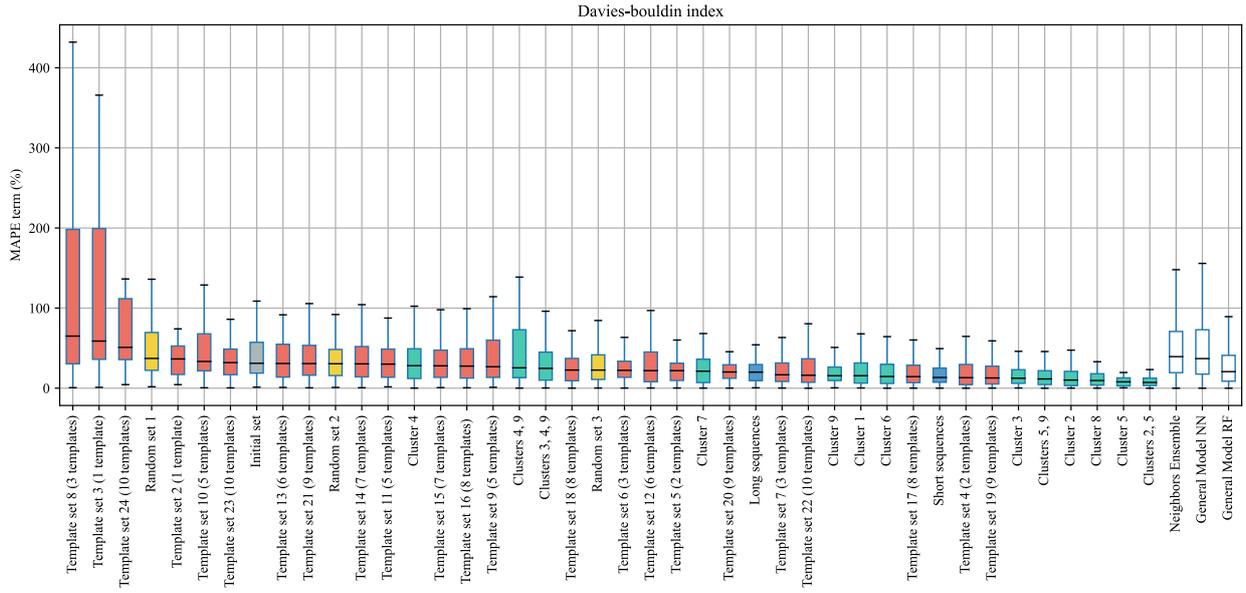

d)

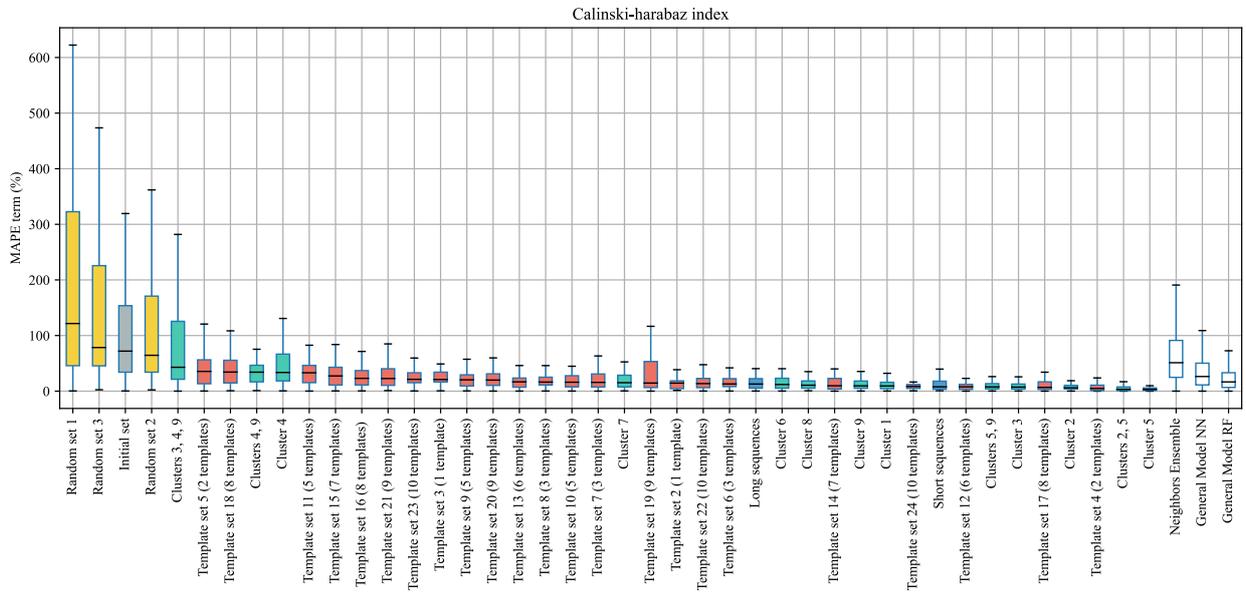

e)

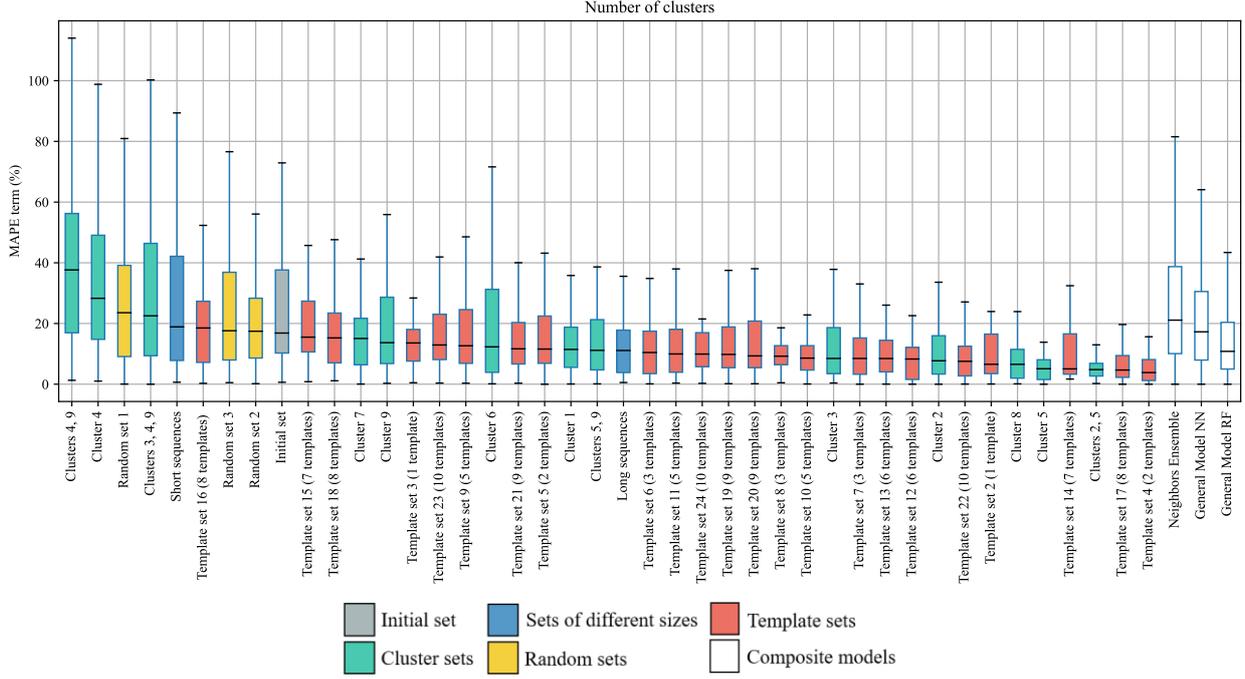

**Figure 9:** Distribution of MAPE terms for the surrogate models for each set (M_each) and composite models (M_gen, M_neig). The boxplot depicts MAPE terms for each target: number of non-clustered sequences (a), execution time (b), Davies-Bouldin index (c), Calinski-Harabaz index (d), Number of clusters (e).

We evaluated the predictive power of the surrogate models for different datasets (see Table 2) to compare the influence of the CP structure, different in various datasets, on the algorithm behavior (see Fig. 9). For non-composite models, the errors are calculated for a random forest trained and tested on a separate set.

The graphs in each figure are sorted by median MAPE, except for composite models, which were moved to the left side of the plot for better comparison with separate set models. Comparing the composite model to the general one (applied to the initial set, as depicted by the gray bar) it can be seen that the General model (RF) mostly shows the best performance (that is also reflected in Table 3).

As can be seen, for some features in separate sets, the following higher error values are the most common: execution time for cluster sets, Davies-Bouldin index for template sets, Calinski-Harabaz for random and cluster sets. At the same time, for template sets, the error of the number of clusters is smaller on average. There are clusters that usually show good (e.g. cluster sets ##2, 5, 8) and bad (e.g. cluster sets ##1, 4, 7, 9) predictive performance. Still, at the same time, there are clusters where predictive power of the surrogate models varies significantly in different quality and performance criteria (e.g. cluster set #3).

Fig. 9 does not include the average ensemble of surrogate models ($M_{aver}$) and outliers for the neural network since their display makes the graph unreadable. Any composite model predicts worse than the best model on a separate set. However, with the exception of the execution time for the k-nearest neighbors ensemble of surrogate models ($M_{neig}$), any composite model on average predicts better than the worst model on separate sets.

# 7. Discussion

The proposed approach can be extended and used in various solutions. This section discusses possible extensions of the approach investigated within the study to show how it could be improved with the integration of additional solutions.

## 7.1. Data collection

One of the important aspects of any machine-learning model (and in particular a surrogate model) is the collection of appropriate data available for training. Within any computational environment, collection of the history of previous runs is essential for better planning of complex computations (see e.g. the concept of provenance [63,64]).. Within metaheuristics, the idea is going further with consideration of history of individuals evaluated through trial and error. Such data collection enables significant improvement of an algorithm, e.g. by avoiding re-evaluation of selected individuals. For example, the LTMA (Long Term Memory Assistance) [65] approach exploits the concept of storing individual evaluations for this purpose.

Within the proposed approach, we focus on the number of cases and domain-specific diversity in training data within knowledge discovery algorithms. In this case, as the prediction depends on data characteristics, the important issue is storing runs on various data sets to cover the data parameter space. Another approach is experimental running of the algorithms on different subsets with random or domain-specific decomposition. The latter can be based on the resulting discovered knowledge (e.g. clustering of the original dataset) to provide better adaptation of the algorithm to case variation observed in data.

## 7.2. Interactive Analysis

Based on the surrogate models mentioned above, analytical instrumental solutions can be developed for interactive analysis of target algorithms. Within our study, we developed a prototype of a software solution for researchers and analysts. The prototype consists of backend and frontend modules. The backend includes a trained general surrogate model with random forests. The frontend is a website with a single page (see Fig. 10) where users can upload a file with new sequences of states. The software makes predictions of the parameters of possible outputs and defines the best ones. Fig. 9a shows a user's opportunities. A user can upload a new file with sequences, select necessary objectives to define better output parameters, and select the type of table visualization (show optimal solutions or all solutions). In the case of two objectives, the scatter plot with solutions is shown under the table with the red-colored best solutions. Such solutions enable investigating the space of input parameters for a new set and deciding which parameters are better for launches of the TDC method. Moreover, users can choose the most crucial output parameters according to their research. For example, some researchers prefer CHI and others prefer DBI to define the best number of clusters. Or, probably, the most important criterion can be time if the system is used in real time.

a)

Upload file with sequences below:

```
Drag and Drop or Select Files
```

The last sequence file was random_set_2.txt

Output Pareto objectives
☑ Davies bouldin score ☑ Non aligned ☐ Time execution

Show:
○ All ● Nondominated only

| № | input increment | input mutation probability | input mutation num | input parent fraction | input start population factor | output calinski harabaz score | output davies bouldin score | output non aligned | output num of clusters | output time execution |
|---|---|---|---|---|---|---|---|---|---|---|
| 1 | 3 | 0.1 | 4 | 0.3 | 1.2 | 29.54 | 4.02 | 28 | 4 | 19.04 |
| 2 | 3 | 0.2 | 4 | 0.3 | 1.2 | 29.45 | 4.2 | 26 | 4 | 20.26 |
| 3 | 3 | 0.2 | 4 | 0.2 | 1.2 | 28.96 | 4.41 | 25 | 4 | 19.55 |
| 4 | 3 | 0.2 | 4 | 0.1 | 1.2 | 27.46 | 4.52 | 25 | 4 | 19.26 |
| 5 | 3 | 0.2 | 2 | 0.2 | 1.2 | 26.68 | 5.22 | 24 | 4 | 18.18 |
| 6 | 3 | 0.2 | 2 | 0.1 | 1.2 | 25.51 | 5.39 | 24 | 5 | 17.97 |
| 7 | 5 | 0.2 | 4 | 0.1 | 1.2 | 24.06 | 5.89 | 24 | 5 | 22.48 |
| 8 | 7 | 0.2 | 4 | 0.1 | 1.2 | 23.91 | 6.19 | 22 | 5 | 23.71 |
| 9 | 7 | 0.2 | 2 | 0.1 | 1.2 | 23.56 | 6.79 | 22 | 4 | 18.44 |
| 10 | 7 | 0.2 | 4 | 0.2 | 1.2 | 17.17 | 6.83 | 21 | 5 | 23.99 |
| 11 | 7 | 0.2 | 2 | 0.2 | 1.2 | 16.45 | 7.22 | 21 | 4 | .94 |

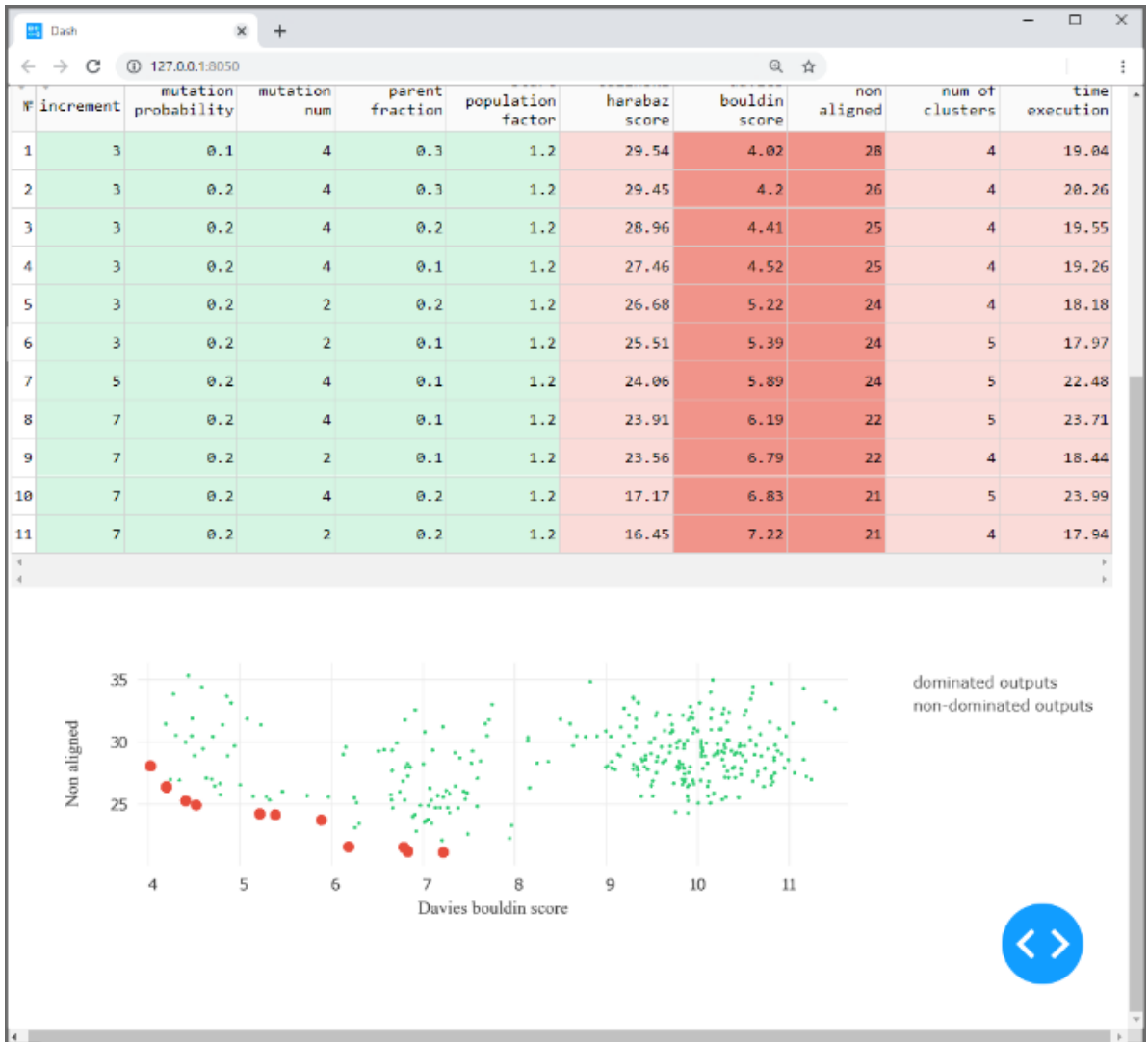

**Figure 10:** Software prototype a) loading data; b) data analysis and visualization

### 7.3. Interpretable Prediction

Fig. 11 depicts the importance of features for the general surrogate model based on RF. For the number of clusters, the parameters of the input set (the number of unique sequences, the number of length outliers, the standard deviation of sequences' length) and the start population factor of the GA are the most important. In contrast, the parameters of states mainly do not influence the number of clusters. For other predicted parameters, some states' parameters are significant. For execution time, the transfers from other departments to cardiology departments or an intensive care unit are important. We assume that the patients move to other departments during their hospitalizations when they have comorbidities (other diseases besides acute coronary syndrome). As a result, their clinical pathways become longer, and the GA works for them longer. The transfer from a coronary catheterization department to a cardiology department has crucial importance for predictions for the Davies-Bouldin index.

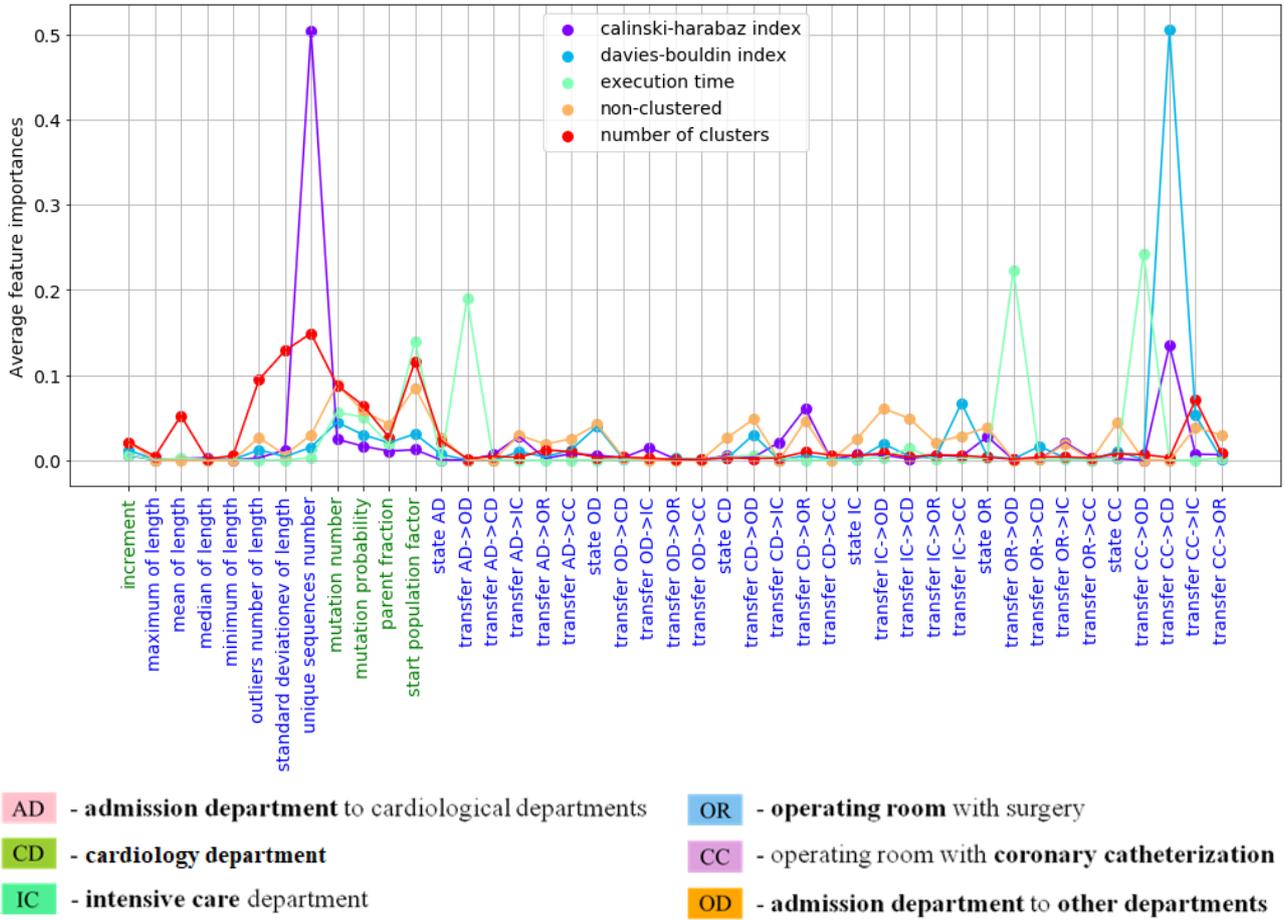

**Figure 11**: Feature importance for the general surrogate model based on RF regression
(the GA parameters are colored in green, the parameters of input sets are colored in blue)

The interpretation above helps to understand how the developed evolutionary algorithms work. It allows one to make a "smart" selection of model parameters and not just go through the hyperparameters in search of the best ones. Such interpretations of model parameters can be the basis for constructing explainable artificial intelligence (XAI) [66]. XAI is aimed to create explainable models and their automatic interpretation. In the future, it is also possible to construct an ensemble of surrogate models that will select the weights of the base models based on their interpretation.

In the future, we plan to optimize the process of tuning surrogate models, develop a version for parallel calculations for big data, and develop a "smart" system for selecting base models and automatic interpretation of the developed models. Also, we consider testing the proposed approach with other problems that our research team solves.

## 8. Conclusion and Future Work

Within the presented study, we investigated an approach for prediction of evolutionary algorithm quality and execution time using surrogate models. Considering the knowledge discovery through the clustering of CPs, we see that the predictive model behaves significantly different in different domain-specific subsets (i.e. different classes of CPs). The best average performance was obtained with random forest model for prediction of all selected characteristics (both in quality metrics and execution time).

The conducted explanatory study through the feature importance analysis shows that the predictive power of a surrogate model is affected by different features for different performance indicators. In general, this can be considered as evidence of the no-free-lunch theorem working in performance prediction. Still, the significance and diversity in the behavior of predictive models and feature importance give the insights that the model can be further developed with extended algorithms of predictive model selection depending on the identified domain-specific features that we consider as one of the potential directions for further development.

Another important direction for further development is application of the proposed approach for algorithm prediction in parameter control and tuning algorithms. With proper selection of the predictive model, surrogate-assisted tuning of various algorithms may be improved. Also, as we have shown, these surrogate models can be the base for decision support systems, analytical, and research solutions. Evolutionary algorithms for multi-objective optimization (MOEA) have been a "hot" topic of research for many years. This is because MOEA is actively used in industrial projects, as it allows modeling complex objects and their dynamics, if necessary [67]. Thus, we believe that the proposed approach can support further development of many applications in various areas.

*Acknowledgments.* This research is financially supported by the Russian Science Foundation, Agreement #19-11-00326.